\documentclass[authoryear]{elsarticle}
\usepackage[margin=2.5cm]{geometry}
\usepackage{amssymb,url}
\usepackage{xcolor}
\journal{Remote Sensing of Environment}

\begin{document}

\begin{frontmatter}

\title{{Deep Learning Models for River Classification at Sub-Meter Resolutions from Multispectral and Panchromatic Commercial Satellite Imagery\footnote{Published under CC-BY-NC-ND license with  doi:10.1016/j.rse.2022.113279}}}

     \author[label1,label2,label5]{Joachim Moortgat}
      \author[label3]{Ziwei Li}
      \author[label1,label2]{Michael Durand}
         \author[label1,label2]{Ian Howat}
        \author[label2]{Bidhyananda Yadav}
                \author[label4]{Chunli Dai}
                
     \address[label1]{School of Earth Sciences,
            The Ohio State University,
             Columbus, 43210, OH, USA}
             
                  \address[label5]{Civil Environmental and Geodetic Engineering,
            The Ohio State University,
             Columbus, 43210, OH, USA}

	 \address[label2]{Byrd Polar and Climate Research Center,
            The Ohio State University,
             Columbus, 43210, OH, USA}
             
              \address[label3]{Computer Science and Engineering,
            The Ohio State University,
             Columbus, 43210, OH, USA}
             
              \address[label4]{Civil and Environmental Engineering,
            The Ohio State University,
             Columbus, 43210, OH, USA}

\begin{abstract}
Remote sensing of the Earth's surface water is critical in a wide range of environmental studies, from evaluating the societal impacts of seasonal droughts and floods to the large-scale implications of climate change. Consequently, a large literature exists on the classification of water from satellite imagery. Yet, previous methods have been limited by 1) the $>10$ m spatial resolution of public satellite imagery, 2) classification schemes that operate at the pixel level, and 3) the need for multiple spectral bands. We advance the state-of-the-art by 1) using commercial satellite imagery with panchromatic and multispectral resolutions of $\sim$ 30 cm and $\sim$ 1.2 m, respectively, 2) developing multiple fully convolutional neural networks ({FCN}) that can learn the morphological features of water bodies in addition to their spectral properties, and 3) {FCN} that can classify water even from panchromatic imagery.  
This study focuses on rivers in the Arctic, using images from the Quickbird-2, WorldView-1, WorldView-2, WorldView-3, and GeoEye satellites. Because no training data are available at such high resolutions, we construct those manually. First, we use the red, green, blue, and near-infrared bands of the 8-band multispectral sensors. Those trained models all achieve excellent precision and recall over 90\% on validation data, aided by on-the-fly preprocessing of the training data specific to satellite imagery. In a novel approach, we then use results from the multispectral model to generate training data for {FCN} that only require panchromatic imagery, of which considerably more is available. Despite the smaller feature space, these models still achieve a precision and recall of over 85\%. We provide our open-source codes and trained model parameters to the remote sensing community, which paves the way to a wide range of environmental hydrology applications at vastly superior accuracies and 1 -- 2 orders of magnitude higher spatial resolution than previously possible.  

\end{abstract}

\begin{highlights}
\item Sub-meter-resolution satellite imagery is used to track river extents
\item Convolutional neural networks can detect water with over 90\% accuracy 
\item Even panchromatic images allow accurate water detection 
\item Multiple {FCN} are adapted for satellite imagery and evaluated for performance 
\end{highlights}


\end{frontmatter}

\section{Introduction}
\label{sec:introduction}
Climate change is causing unprecedented alterations in the Earth's surface hydrology such as the melting of Arctic ice and alpine glaciers, more extreme flooding and droughts in semi-arid zones,  variations in precipitation more generally, and rising ocean levels. Monitoring these impacts is challenging, particularly in geographically remote and economically under-resourced regions \citep{shiklomanov2002widespread,hannah2011large,unganai1998drought,rouault2005intensity,pope1992identification}. 

Remote sensing is an increasingly affordable and scalable alternative to ground measurements, and indeed global surface water maps have been generated from, e.g., Landsat multispectral imagery \citep{pekel2016high}. Hundreds of publications are devoted to the remote sensing of water in particular study areas (e.g., for rivers in \cite{du2012estimating,ottinger2013monitoring}), as well as a plethora of techniques to classify water from multispectral satellite data \citep{huang2018detecting}. Approaches based on current technology have enabled regional and global-scale advances in measuring rivers and characterizing the global fluvial system  \citep{allen2015patterns,pekel2016high,Allen2018, Feng2021}. 

However, there are two areas in which the remote sensing of surface water {from satellite imagery} can be significantly improved, which is the objective of this work. 

First, most prior research has relied on publicly available satellite imagery with multispectral resolutions of tens of meters (e.g., Landsat and Sentinel-2). While those satellites offer excellent spatiotemporal coverage, only wide rivers ($\gtrsim 90$ m {for 3 Landsat pixels or $\gtrsim 30$ m for 3 Sentinel-2 pixels}) can be resolved and only large \textit{changes} in water bodies can be tracked over time. To track smaller rivers and changes in water bodies with current satellites, one has to rely on commercial imagery at the expense of spatiotemporal coverage. In this work we use imagery from the GeoEye, Quickbird-2 (QB2), WorldView-1 (WV1), WorldView-2 (WV2), and WorldView-3 (WV3) satellites which have multispectral and panchromatic resolutions of around 1.2 m and 30 cm, respectively. { We acknowledge the limitations of using commercial imagery, but currently this is the only platform that combines such meter-scale resolutions with repeat coverage over much of the Arctics. High spatial resolutions can of course also be achieved with airborne platforms, but at even lower spatiotemporal coverage.  }

Second, the majority of widely used water classification techniques only use spectral information at the pixel-level, or sometimes a small kernel around each pixel \citep{dai2018estimating}. As a well-known example, the (modified) Normalized Difference Water Index (NDWI, \cite{gao1996ndwi}) compares the difference between intensities in the green and near-infrared bands (or a middle infrared band in MNDWI, \cite{xu2006modification}). An `Automated Water Extraction Index' was proposed in \cite{feyisa2014automated} and compared to other single- and two-band pixel-level classifiers. Other prominent works (e.g., \cite{pekel2016high}) use `expert systems' that involve complex sequences of finely tuned conditional statements. In recent years, various machine learning techniques, such as random forest classifiers, have also been used to learn more complex relationships between, e.g. Landsat's top-of-atmosphere (TOA) reflectance and water indices \citep{ko2015classification}. 

All the aforementioned types of classifiers suffer from the same fundamental limitation, which is that there is simply only so much information contained in a single pixel (and perhaps its neighbors). Pixel-level classifiers often have a reasonably high recall, meaning that they can correctly identify a large fraction of water pixels, but a low precision, i.e., misclassifying many features as water (such as roads, shadows, buildings, clouds, etc.). Manual tweaking of parameters may be required to find an optimal balance between recall and precision, together with post-processing. As we demonstrate below, this process is time consuming, subjective, and not scalable. 

At the same time, the pixel-level classification from images, referred to as semantic segmentation, is a common computer vision task. In recent years, increasingly powerful techniques have emerged from the field of deep learning, or more specifically, fully convolutional neural networks ({FCN}, \cite{long2015fully}). These neural networks cannot only learn highly non-linear relationships between all available spectral information at the pixel level but also detect large-scale features such as edges and gradients, and thus morphologies. For these reasons, in recent years, FCN have started to be adopted for land-use classification in hydrology and other Earth Science disciplines, using in situ cameras \citep{eltner2021using}, airborne platforms \citep{carbonneau2020adopting,weng2018land,liu2018object}, and multispectral \cite{isikdogan2019seeing} and hyperspectral \citep{qin2021small} satellite imagery. 

We {expect} that these capabilities will allow {FCN} to outperform most if not all pixel-level classification schemes for satellite imagery, whether to detect water or other features. 

{
The goals of this work are as follows:
\begin{enumerate}
\item Adapt and optimize several of the most successful recent FCN architectures in computer vision \citep{ronneberger2015u,chaurasia2017linknet,he2016deep,isikdogan2019seeing} for satellite remote sensing, e.g.~to allow for exceedingly large image sizes (easily $>10^9$ pixels) and any number of spectral bands; use satellite instrument specific noise terms in model training to improve robustness.
\item Generate high-quality training data at meter-scale resolutions by labeling each pixel as either water or land in hundreds of full-size satellite images. As is common in supervised machine learning, this is the most time-consuming and labor intensive component of the work. 
\item Train all FCN models on the aforementioned training data and rigorously evaluate the performance of each model in terms of accuracy, computational efficiency, and memory requirements. A hold-out set of imagery from geographically distinct locations is used to assess the generalizability of predictions. 
\item In a novel approach, we use water labels derived from multispectral images to generate training data for models that only use \textit{panchromatic} information. The archive of panchromatic imagery is considerably larger and has a higher resolution. Multispectral images can also be sharpened to the panchromatic resolution, and combinations of panchromatic and multispectral sensors will likely be adopted broadly in the future \citep{durand2021achieving}. 
\end{enumerate}}

 To bridge the temporal and spatial resolution gaps between commercial versus public imagery, in a future work we will also explore \textit{super-resolution} FCN to achieve meter-scale resolutions from 10-m Sentinel-2 imagery, which has a 5-day temporal resolution.

\section{Methods}
\label{sec:methods}

\subsection{Imagery}
\label{ssec:ims}
Imagery from the GeoEye, QB2, WV1, WV2, and WV3 satellites was generously provided by the Polar Geospatial Center at the University of Minnesota, orthorectified and in the Sea Ice Polar Stereographic North coordinate reference system (ESPG:3413).  All multispectral images have red, green, blue (RGB), and near-infrared (NIR) bands. Roughly half of the multispectral images have an additional 4 bands, which are coastal (C, green), yellow (Y), another red band (RE), and another infrared band (N2). All bands are provided at the same resolution of around 1.2 m. 

Each multispectral image has a synchronous panchromatic counterpart from a different sensor on the same satellite, while many more panchromatic images are available without a multispectral counterpart either because of data transfer constraints or the lack of a multispectral sensor on, e.g., the WV1 satellite. The panchromatic images have resolutions between 30 -- 50 cm.  
Our imagery covers all of the Arctic in Alaska and Canada and over a 16 year time period from 2004 to 2020. 

Images are of the order of $\sim 12 \times 12\ \mathrm{km}^2 $ or 10,000 $\times$ 10,000 pixels at multispectral resolutions (and 16 times higher for panchromatic).  Light intensity values for all bands are recorded as 11-bit integer Digital Numbers. 

{
\subsection{Machine learning terminology and concepts}
\label{sec::terminology}
Machine, or deep, learning algorithms have typically first been developed in computer science and data analytics communities that use different terminologies from what some readers of this journal may be familiar with. It may be useful to define in this section the translation between some important terms to provide rigorous definitions in the context of deep learning, while allowing the use more familiar terms in the following sections. 

\subsubsection{{Machine learning as error minimization problem}}
Training a machine learning model is equivalent to `fitting' a (linear or nonlinear) model to a number of measurements of the variables/features of interest in a numerical regression. The best fit is defined by the smallest fitting error, which is referred to as the `loss function' in machine learning. Different loss functions can be used, such as absolute or mean squared errors. We will use a so-called `cross-entropy' loss function, which is essentially just a mean squared error but generalized for classification (`logistical regression'). 

There are many ways to minimize a fitting error. The most common in machine learning is the `gradient descent' method. It starts with an initial guess for fitting parameters and then evaluates not only the fitting error, but also the \textit{derivative} of the error. For the next iteration, fitting parameters are adjusted by a small step in the downward direction of the error slope, in order to move closer to the minimum error. The size of this step is the `learning rate'. Larger learning rates could converge to the best fit (minimum error) faster, but risk over-shooting. Small steps are safer but slower. A common approach, which we adopt, is to use an adaptive learning rate that starts out large but is automatically decreased as the minimum error is approached. 

\subsubsection{{Image segmentation}}
The closely related terms of image segmentation, or semantic segmentation, or instance segmentation refer to the task of classifying each pixel in an image (e.g., as water, land, forest, etc.). In FCN, the classification of one pixel does not just depend on the panchromatic or multispectral intensities of just that pixel, but on those of all the pixels in a large window around that pixel. This sphere of influence is the `receptive field' of the model. This is somewhat equivalent to the number of variables, or features, in a numerical regression and can easily be of the order of $10^5$. 

Because we want to fit FCN to large numbers of satellite images, each of which has of the order of $10^8$ pixels, and the FCN models have millions of trainable parameters, it is not possible to use the gradient descent method directly on all training data at once, even on the most powerful current GPUs. Instead, it is common to use a stochastic approach in which the model is either fit to a single image at a time (`stochastic gradient descent') or to a small batch of, say, 24 or 32, images (`mini batch gradient descent'). In one `epoch', the FCN is fitted to random batches of images until it has seen all images, with each batch producing different fitting parameters. Next, this process is repeated for a few dozen epochs until the fitting parameters and associated errors converge to a global optimum across all images. 

\subsubsection{{Model training on satellite imagery}}
The aforementioned approach is sufficient for widely used benchmark databases that consist of relatively small labeled RGB images, of the order of $256 \times 256$ or maybe $512\times 512$ pixels.  
However, for $10^8$ pixel satellite images, even a batch size of one is not computationally feasible. To \textit{train} a FCN on satellite imagery, it is necessary to chop the images into smaller tiles. A single full-size image can provide several hundred such training tiles, and by randomly selecting small batches of tiles across all training images, the fitting process is more robust and efficient. Because a FCN does not `see' more than the receptive field around each pixel, there is no information loss in training on tiles that are at least the size of the receptive field.

Training on smaller tiles also offers another advantage. In a typical full size satellite image, a river may only occupy a small fraction of the pixels. In other words, there is a significant imbalance between the land and water labels, and this negatively impacts convergence of the training process. When we tile the full-size images, we discard the tiles with less than 0.1\% water labels, such that each tile has at least a few river pixels. 

Once the model is trained, the `training weights' are equivalent to the best fitting parameters of a (linear or nonlinear) numerical regression and fully specify the final FCN model. We can then use these models to make \textit{predictions} of, e.g. river classification, on new images. In machine learning, this is referred to as `inference'. 

Importantly, these predictions are made for one image at a time and we will discuss below how our most memory-efficient FCN can do so on full-size images using even consumer-grade GPUs as well as CPUs (the training process is far more computationally and memory expensive than the inference).

\subsubsection{{Model accuracy}}
The last concept to discuss is the metrics used to evaluate the performance of different FCN. The loss function is one metric, but it is not particularly insightful. Instead, we use the more intuitive 
 precision, recall, and F1 scores (see, e.g., \citet{james2021convolutional}). Conceptually, precision is the percentage of pixels that are predicted to be water and were also labeled as such in the training data. Recall is the percentage of pixels labeled as water that were also identified as such by the model. The F1 score is the harmonic average of those two metrics, which is close to the worst of those two numbers. 

As two extreme examples: if a model predicts that only one pixel of an image is water and that pixel is indeed labeled as water, it would have a precision of 100\% but recall of nearly 0\%. Conversely, if the model predicts that all pixels are water, it would have a recall of 100\% but a low precision. In both cases, the F1 score is low. Only if both precision and recall are high is the F1 score high, which is why F1 is a useful single metric to represent classification `accuracy' (see also \citet{carbonneau2020adopting}). 
 }
 
\subsection{Training data for multispectral images }
\label{ssec:training}
		 \begin{figure}[h!]
 \noindent\centerline{\includegraphics[width=\textwidth]{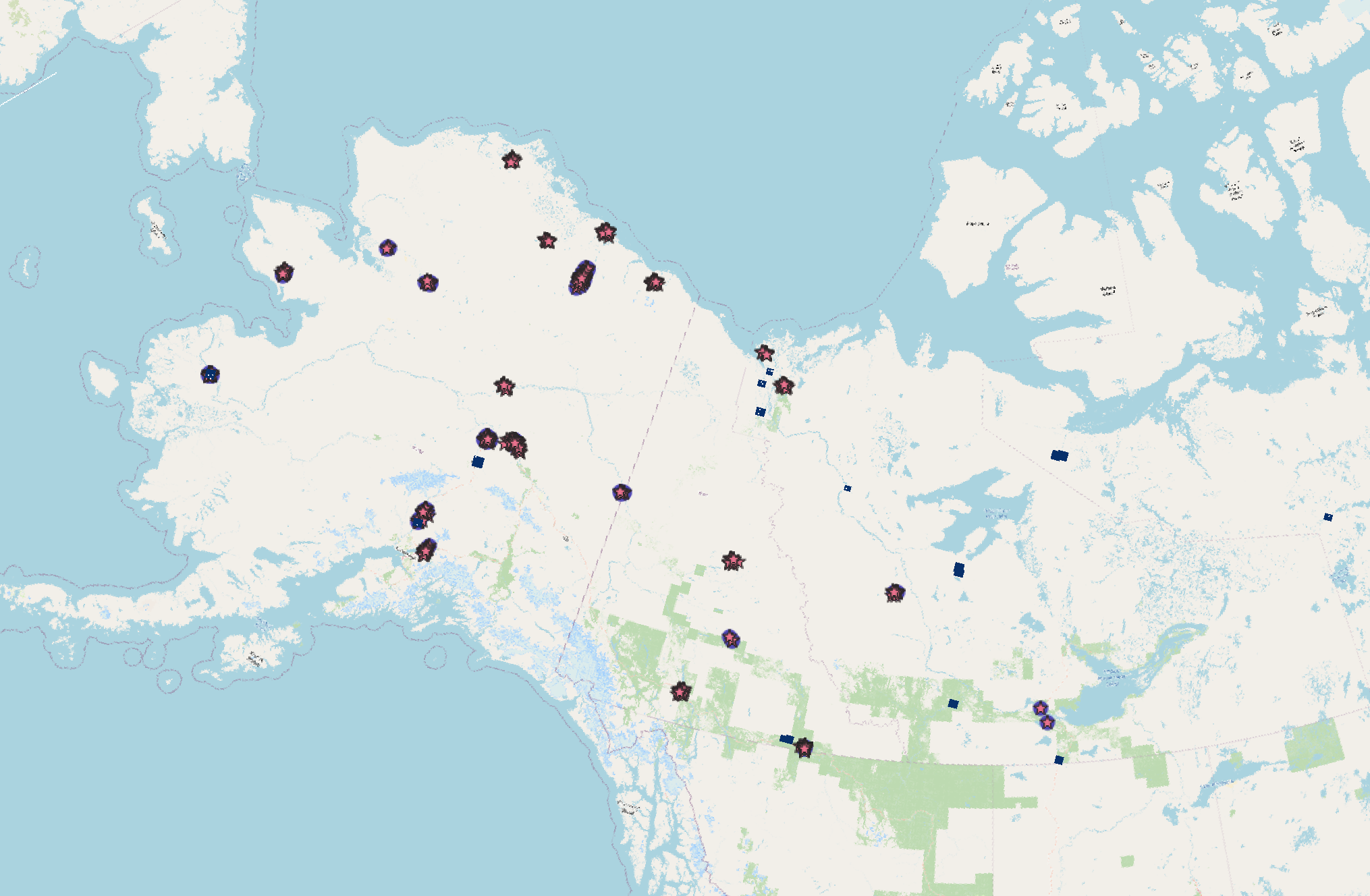}}
\caption{Locations of training and validation data. {Circles and stars are for the 125 and 1,150 full-size images used in the first and second iterations of model training, respectively. The squares show the (actual size of) independent full-size images used for additional validation.}
 \copyright~OpenStreetMap contributors.
\label{fig::fig0}}
\end{figure}
To the best of our knowledge, no suitable labeled training data for water detection exists at centimeter to meter resolutions. Using classifications from earlier works based on, e.g., Landsat and Sentinel imagery would result in poor quality training data because those images (at 10 -- 30 m resolutions) cannot resolve smaller rivers, islands, braids, and sandbanks. 
As is often the case in (supervised) machine learning, the most time-consuming component of our work was to manually construct our own training data. 

\begin{figure}[h!]
 \noindent\centerline{\includegraphics[width=\textwidth]{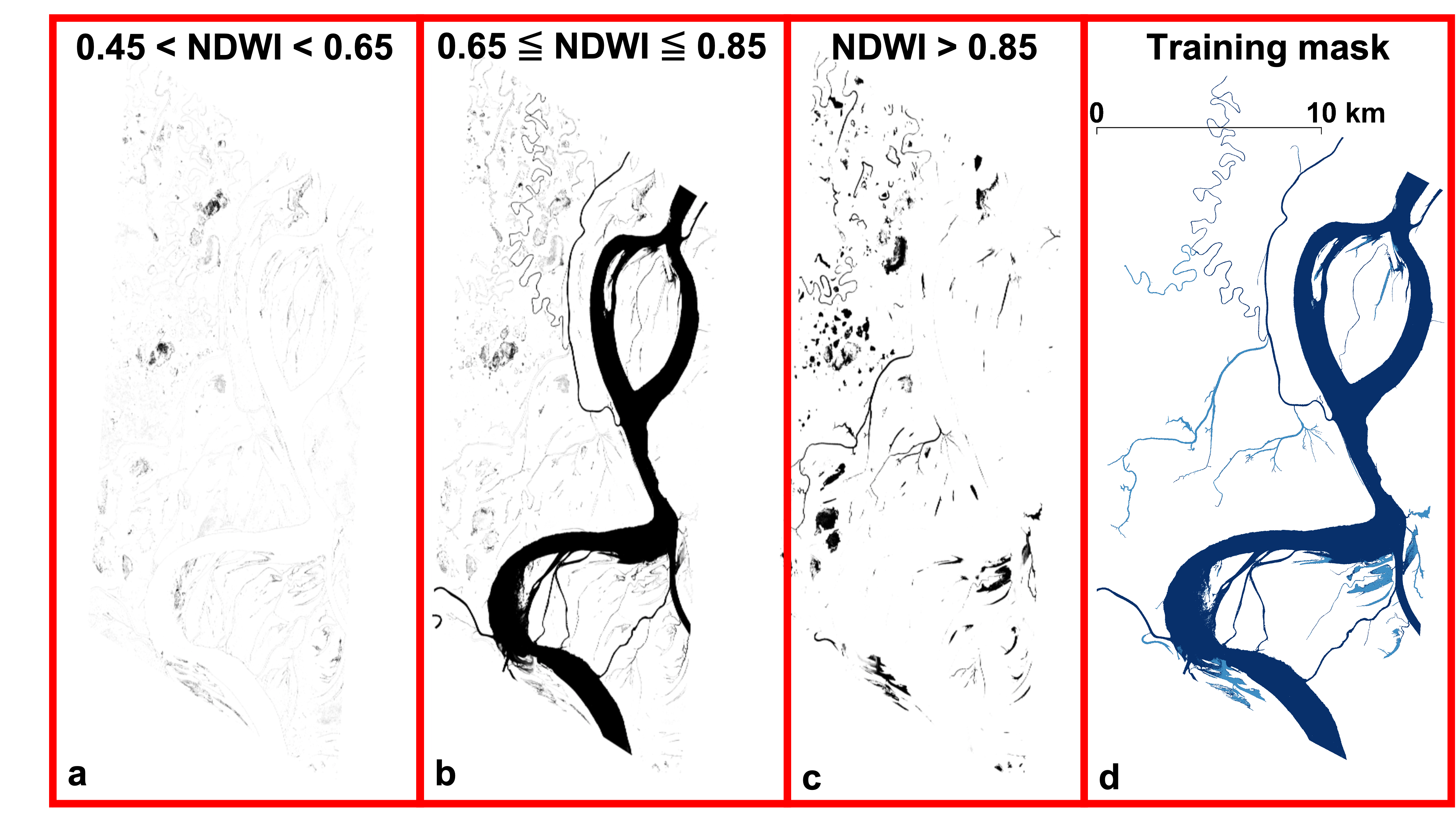}}
\caption{Construction of water training labels for a single WV3 image covering a section of the Yukon. Pixels are binned based on manually tuned NDWI thresholds (\textbf{a}--\textbf{c}) 
 together with  post-processing steps to obtain final training labels  (\textbf{d}).
\label{fig::fig1}}
\end{figure}

We adopted an iterative approach to construct a sufficiently large set of labeled images. In the first iteration, we selected 125 mostly cloud-free full-size multispectral images from WV2, WV3 and QB2 covering 15 different river sections throughout Alaska and Canada (Figure~\ref{fig::fig0}). Our aim is to label each pixel in these images as either land or water (semantic segmentation).  
To be more precise, we allow for ‘soft labels’ which are a continuous scale of probabilities between 0 (land) and water (1){, or rather, for optimal storage, we use 8-bit integer labels between 0 and 255.

Our manual labeling of these images proceeds along the following steps:
\begin{enumerate}
\item Apply a top-of-atmosphere (TOA) correction to correct for the different path lengths that light has to travel through the atmosphere for different satellite orientations, recorded in the imagery metadata. 
\item Normalize data to account for different ranges in Digital Numbers. 
\item Use (modified) NDWI thresholding to bin pixels into likely-land and likely-water. For 8-band WV images, the coastal (C) and far infrared (N2) bands provide better land-water contrast.  Traditionally, a single threshold on NDWI is used to establish a binary mask of, for example, land for NDWI $<$ 0.22 and water for NDWI $\geq$ 0.22 \citep{mateo2021towards}. 
In our work, we manually tune \textit{three} threshold values TH1, TH2, TH3. Pixels with NDWI$<$TH1 are labeled as 0 (land), TH1$<$NDWI$<$TH2 is labeled as 70 (maybe water), TH2$<$NDWI$<$TH3 is labeled as 170 (likely water) and NDWI$>$TH3 is labeled as 255 (definitely water). The threshold values are determined by visual inspection of the labels overlaid on the associated true color images. Optimal thresholds differ significantly between different images, even after applying the TOA correction, but are of the order TH1=0.3, TH2=0.5, and TH3=0.7. For some images, the highest NDWI values are clearly not water and TH2$<$NDWI$<$TH3 is labeled as 255. This is illustrated in Figure~\ref{fig::fig1} for a section of the Yukon river. The leftmost panel for 0.45$<$NDWI$<$0.65 contains mostly pixels that should be classified as land, but also still clearly shows the river shoreline which could be labeled as ‘maybe water’. The middle panel for 0.65$<$NDWI$<$0.85 contains all the pixels that are ‘definitely water’ but also considerable noise, while the rightmost panel for NDWI$>$0.85 has mostly pixels that should be labeled as land but also includes some shoreline features as well as smaller river branches that could be labeled as ‘probably water’. 
\item Even for optimally tuned NDWI ranges, this pixel-level classification exhibits a large fraction of misclassified pixels (easily over 30\%). To eliminate noise away from the main river body, we perform a  largest-connected-component analysis (LCCA). The idea being that the main river body should be the largest connected region of pixels labeled as water. However, there may be some gaps in the classification of the river body, especially for small rivers or river branches. Reasons could be small clouds, shadows and, at meter-scale resolution, even bridges.  Before performing the LCCA, we therefore need to make sure that all the river pixels are connected. To do so, we create a separate binary mask in two steps: 1) we take advantage of \textit{a priori} river centerlines from the SWORD database \citep{altenau2021surface} and overlay this on the NDWI mask, and 2)
we apply a Gaussian blur kernel to the NDWI classification to define a region somewhat wider than the actual river body (and also enlarging other mislabeled pixel clusters). This auxiliary mask is made binary by setting all non-zero values to one (the NDWI mask itself is not altered in this step).
\item We apply the LCCA to the binary auxiliary mask and only keep the largest component. This eliminates all disconnected clusters of pixels mislabeled as water, such as lakes, buildings, and clouds. We then multiply this auxiliary mask with the NDWI mask. All pixels that were labeled as land by NDWI thresholding remain so, while all pixels labeled as water by NDWI that are not part of the largest component are multiplied by zero and thus correctly relabeled as not-river. 
\item A denoising step is performed by combining erosions and dilations to eliminate individual mislabeled pixels within the river body. 
\item Some features like a cloud, shadow or nearby road cannot be automatically separated from the river and have to be manually removed. 
\end{enumerate}  
Figure~\ref{fig::fig1}d shows an example of the final labels for the Yukon.

The aforementioned process of generating training data illustrates the limitations of using (M)NDWI or other pixel-level classifiers. It is difficult, if not impossible, to generalize and automate, while {manual} tuning is time-consuming and not scalable. 
Labeling 125 full-size images by this approach required around 3 weeks of full-time work (roughly an hour per image). 

In the context of deep learning, though, 125 training images is not a lot and we aimed for an order of magnitude more. Doing so with the above process is not feasible, so we already deploy FCN to more efficiently generate training data. Specifically, we first trained our FCN on the 125 manually labeled images and then used the trained FCN to classify thousands of other full size images. We then visually selected the 1,150 best masks (with locations shown in Figure~\ref{fig::fig0}) and performed the same post-processing steps to obtain high qualify labeled images. All 1,150 masks were overlaid on their corresponding images for quality control. While still time-consuming, this process is far more efficient than the labeling of the initial 125 images. 

Both sets of full size training images were merged with the labels and tiled into smaller 732 $\times$ 732 pixel chips for reasons discussed earlier, eliminating chips without any water pixels. This resulted in 4,606 labeled image tiles in the first set of training data and around 14,000 in the second iteration. 
 }
 
 		 \begin{figure}[h!]
 \noindent\centerline{\includegraphics[width=\textwidth]{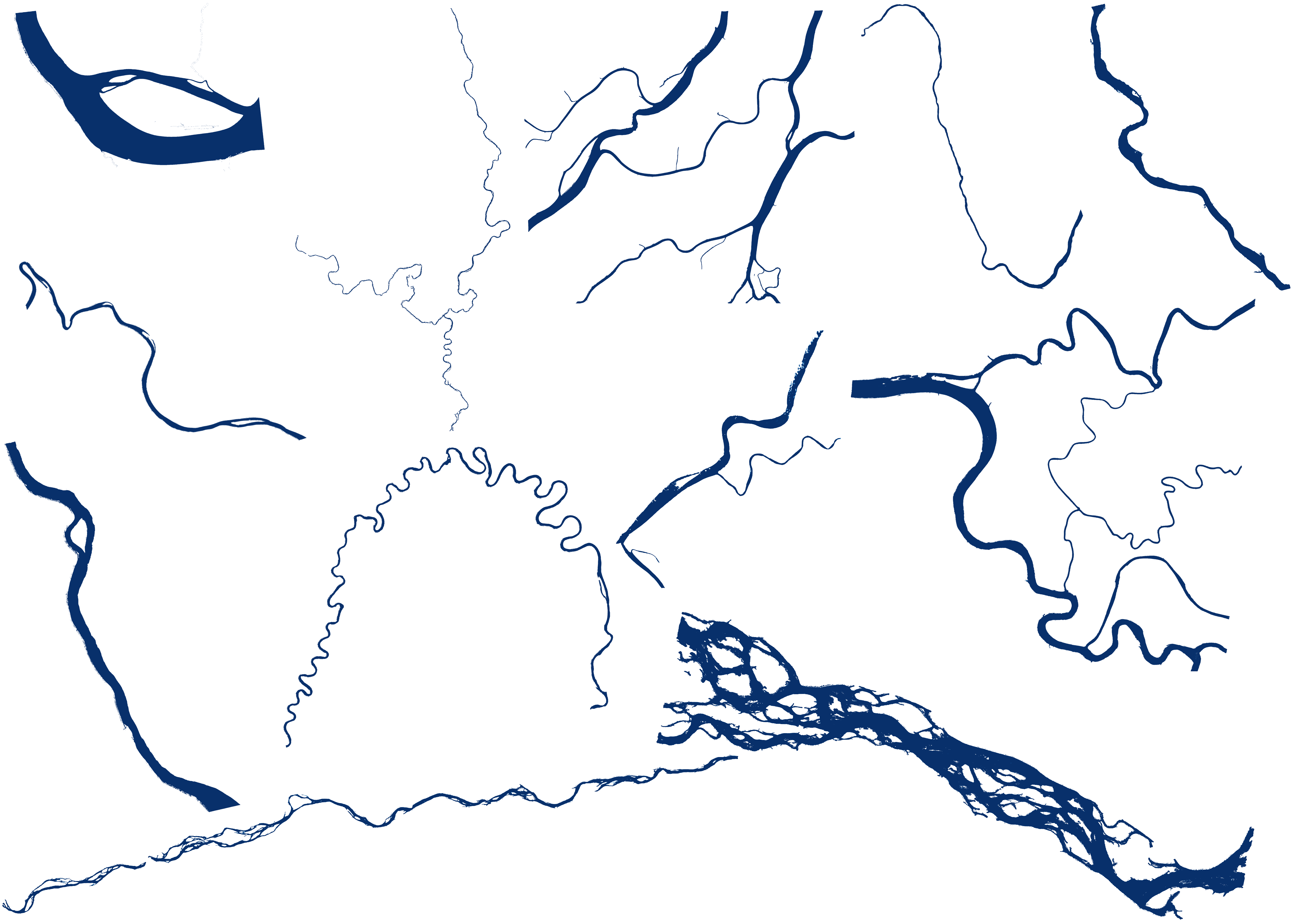}}
\caption{River labels for 12 full-size images in locations distinct from the training data.}
 \copyright~OpenStreetMap contributors.
\label{fig::figx}
\end{figure}

Both labeled image sets are split into 80\% for training and 20\% for validation and stored as {binary compressed} Tensorflow records to reduce storage requirements and improve computational efficiency {(Tensorflow is Google's open source machine learning framework).}  

{As a more robust additional model validation, we manually labeled another 17 full-size images in locations that are geographically distinct from the training data sites (Figure~\ref{fig::fig0}). These full-size images ($>10,000\times 10,000$ pixels) are equivalent to $>8,000$ image tiles of the size used in training the models, but we will perform classification on the full-size images, which therefore have a larger imbalance in the river versus land pixels. Figure~\ref{fig::figx} shows 12 of these labeled images as an illustration of the different river morphologies in this set, which contains a mix of narrow and wide rivers with single or multiple branches and different levels of sediments.}

\subsection{Training data for panchromatic images }
\label{ssec:pantraining}

As mentioned in the Introduction, classifying water from single-band panchromatic images is more difficult than from multispectral data that are more information-rich. Even constructing training data  is challenging.  

We propose a novel approach to generate training data for models that can classify water from panchromatic images. Each multispectral image has a panchromatic counterpart (but not vice versa), so we can use the labels derived from the multispectral information for those corresponding panchromatic images. {We can either up-sample the labels to the four times higher panchromatic resolution or down-sample the panchromatic images to the $\sim 1$ m multispectral resolution. For computational and memory efficiency, we choose the latter. }
To be specific: for each labeled image in the multispectral training datasets we find the corresponding panchromatic image, perform a bilinear down-sampling to the multispectral resolution, merge the labels derived from the multispectral information with the panchromatic image, and tile into the same size chips as before.  

Through this fortuitous availability of multispectral-panchromatic image pairs, our training data for any panchromatic {FCN} classifier should theoretically be as good as for models that use full multispectral information.  

We note that all trained models can still perform \textit{predictions} on the highest $\sim 30$ cm resolution images, as shown in the Results section.  {Notably, we also test the panchromatic model predictions on the 17 hold-out full-size images discussed in the previous section.}

 \subsection{Deep learning models }
\label{ssec:models}
While acknowledging the fast evolving field of deep learning, and artificial intelligence more broadly, we consider what appears to be the latest state-of-the-art in computer vision at the time of  this research.  A family of so-called `U-NET' fully convolutional neural networks has proven to be extremely powerful for the task of accurate but computationally efficient image segmentation. U-NET refers to the shape of deep neural networks with several downscaling or encoding layers that, roughly speaking, learn features at increasingly coarse scales, followed by an equal number of upscaling or decoding layers back to the original image resolution. 

The original U-NET model was developed for biomedical image segmentation \citep{ronneberger2015u}. Since then, many variations have been proposed \citep{zhou2018unet++,huang2020unet,zhou2018d} with different numbers and types of encoders, also referred to as `backbones'. In this work, we consider the popular 18- and 34-layer \textit{residual networks} as backbones \citep{he2016deep}. We will refer to these combinations of U-Net with ResNet backbones as U18 and U34. 

Two years after the introduction of U-NET, LinkNet \citep{chaurasia2017linknet} was proposed as a an alternative that focusses more on computational efficiency than accuracy. Like U-Net, different backbones can be used with the LinkNet architecture and again we evaluated ResNet-18 and ResNet-34 (together referred to as L18 and L34). 

While the various U-Net and LinkNet flavors are general purpose image segmentation tools that have been used to identify a wide range of features in RGB photographic images and videos, another {FCN} was developed specifically to classify water from multispectral satellite imagery. DeepWaterMap 2 (\cite{isikdogan2019seeing}, referred to below as DWM) uses the RGB, NIR, and two SWIR bands of Landsat-8 images. That model was trained on the Global Surface Water dataset \citep{pekel2016high}, which is readily available in the Google Earth Engine \citep{gorelick2017google}. It achieved an impressive precision and recall of 97\% and 90\%, respectively, across that entire dataset. 

We modified each of the aforementioned {five models (U18, U34, L18, L34, and DWM)} to work optimally for multispectral and panchromatic satellite imagery. This meant 
\begin{enumerate}
\item including an input layer that allows for any image size, accommodating our large $\geq 10^8$-pixel full-size satellite images, \item allowing for any number of spectral bands rather than just the typical three RGB channels, \item maintaining the geo-referenced information in the geotiffs, and \item several preprocessing steps specifically designed for satellite imagery at the model training stage as described further in the next section. 
\end{enumerate}

{
We trained each model using a {mini batch gradient descent method with batches of 24 images and adaptive momentum optimization of the learning rate to minimize a standard entropy loss function (concepts discussed in Section~\ref{sec::terminology})}. All models were set up to train for 100 epochs but converged in less than 50, requiring only a few hours of computation on a single NVIDIA RTX 3090 or A40 GPU. }

 \subsection{Data pre-processing}
\label{ssec:preprocess}

A number of on-the-fly preprocessing steps are used as a form of data augmentation during the training of all our models (based on \citet{isikdogan2019seeing}). First, we randomly crop 512 $\times$ 512 pixels out of the 732 $\times$ 732 labeled training images. This chip size allows for multiple two-fold spatial scale reductions, one for each encoder layer.  

To mimic the spectral response of typical satellite sensors, information is leaked between neighboring spectral bands in a Gaussian fashion for the multispectral images, followed by random additional Gaussian noise \citep{isikdogan2019seeing}. Finally, `min-max' scaling is applied to both the image digital numbers and the labels, as is typical in machine learning.  

These data augmentation techniques make trained models, even when using relatively small numbers of training images, robust in recognizing similar noise and spectral responses when performing classification on massive amounts of full-size images, and even when using images from different satellites as we demonstrate below. {No top-of-atmosphere corrections are used for the training of our model, nor when making classification predictions.}

{
 \subsection{Data post-processing}
When performing inference (classification predictions), a soft threshold is applied to the FCN output layer to separate predictions into likely-land and likely-water. Specifically, if we denote the initial FCN predicted labels as $x$, we threshold those values by a sigmoid-like function $$\tilde{x} = \frac{1}{1+\mathrm{e}^{-16 (x-0.5)}}$$ as in \citet{isikdogan2019seeing}. To obtain true \textit{binary} land-water masks, we can apply an additional hard threshold at, e.g., $\tilde{x}=0.5$.
}

\section{Results}
\label{sec:results}

\subsection{Training results for multispectral imagery and models}
\label{ssec:multiresults}
Table~\ref{table::metrics} summarizes the precision, recall, and F1 accuracy metrics as well as the GPU training time for all models developed for 4-band imagery. Results are only shown for the second iteration of training data{, i.e.~for the set of 1,150 full-size labeled images}. The metrics for the first iteration of training data{, using only 125 images,} are within 0.5\% of those in Table~\ref{table::metrics} while the GPU cost is proportional to the number of training images.

\begin{table}[h!]
\begin{centering}
\begin{tabular}{|c|cc|cc|cc|c|}\hline
Model & \multicolumn{2}{c|}{Precision  \% } & \multicolumn{2}{c|}{Recall  \% }   &  \multicolumn{2}{c|}{F1  \% } & GPU time \\
& train& val & train& val & train & val  & (epoch 80) \\\hline
DWM  		&	91.4 & 90.6	& 90.9 & 90.5	& 91.1 & 90.4	& 3h 31m \\
U18  		& 93.2 	& 92.5 &	92.0 & 89.9	& 92.6 & 91.1&	3h 30m\\
U34  		& 93.4 	& 92.4 &	92.3 & 90.5	& 92.8 & 91.7	& 4h 11m \\
L18 		&	92.8 & 92.4	& 91.7 & 89.7	& 92.2 & 90.9	& 2h 49m \\
L34  & 93.3 	& 92.3 	&	92.1 & 90.0 &	92.6 & 91.0	& 3h 29m \\\hline
\end{tabular}
\caption{Summary of training results for five different {FCN} applied to $\sim 14,000$ 4-band (RGB-NIR) multispectral images}
\label{table::metrics}
\end{centering}
\end{table}

		 \begin{figure}[h!]
 \noindent\centerline{\includegraphics[width=\textwidth]{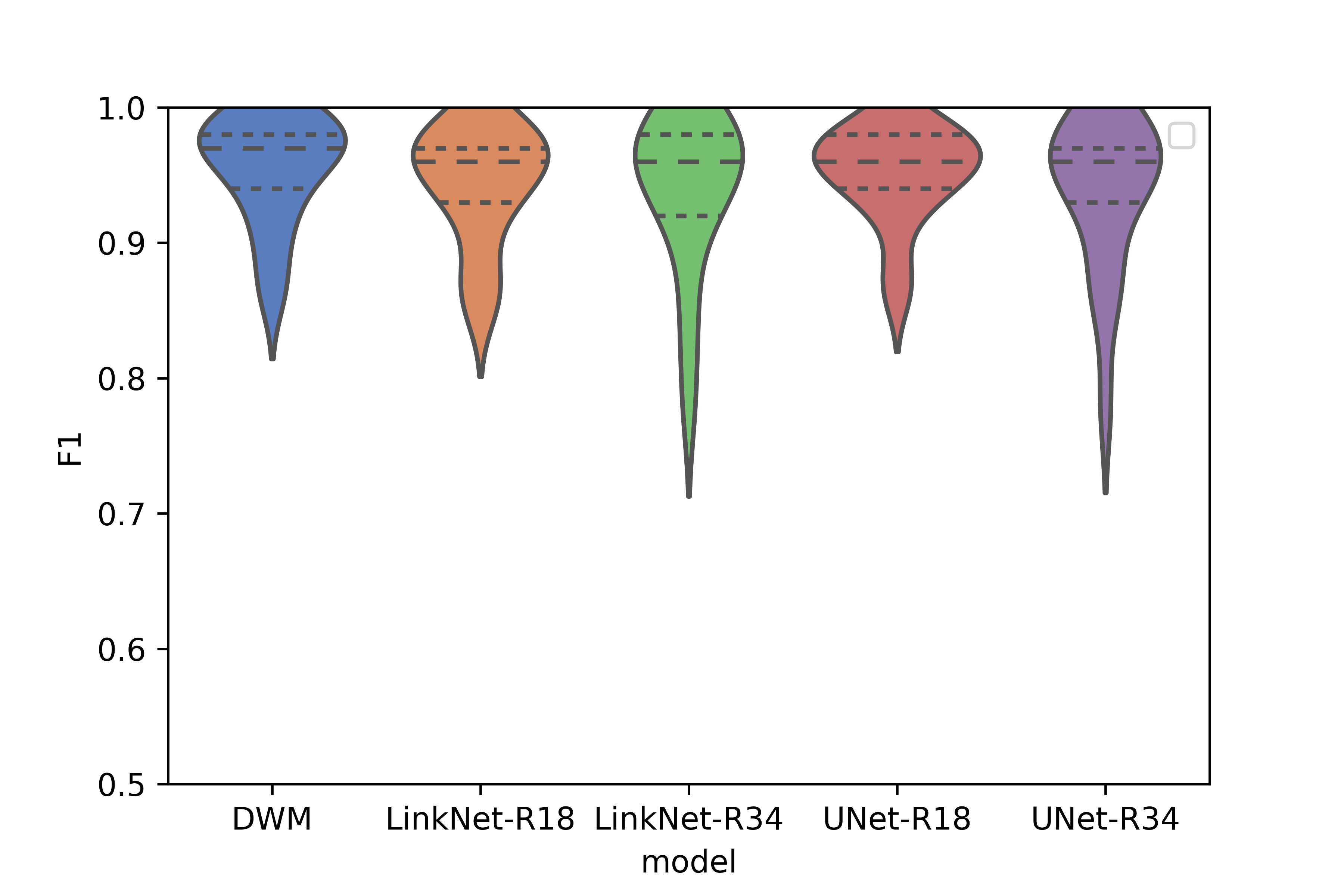}}
\caption{Violin plot for the F1 classification accuracy of all 5 \textbf{multi-spectral} models for 17 independent full-size images. F1 histogram shown in color, short dashed lines show upper and lower quartiles and long-dashed line is the median.}
\label{fig::figx2}
\end{figure}

All {FCN} models show excellent (and similar) performance in terms of these training and validation metrics as well as computational costs. Linknet is faster than U-Net, as was the motivation of its development \citep{chaurasia2017linknet}. The training time for L34 is similar to that of DWM. 

{
When we use the trained models to classify the 17 full-size hold-out images in different geographical locations, we find the F1 accuracy metrics shown as violin plots in Figure~\ref{fig::figx2}. Interestingly, the average F1 score across the 17 images,  $>94\%$, is even higher than for the training and validation data. Most likely, the quality of the images and labels in this test set are better than the average of the 1,150 full-size images used for training. }

\subsection{Training results for panchromatic imagery and models}
Panchromatic images contain significantly less information than the increasing number of (hyper-) spectral bands in new generations of satellites. Yet, there is a large archive of sub-meter resolution panchromatic data, which also goes back further in history (e.g., WV1 only had a panchromatic sensor). For these reasons, we present FCN tools for water classification solely from 1-band images.

Presumably because of the lower information density, these panchromatic models appear to benefit more from our larger second iteration of training data, for which Table~\ref{table::metricspan} summarizes the accuracy and computational cost. 

\label{ssec:panresults}
\begin{table}[h!]
\begin{centering}
\begin{tabular}{|c|cc|cc|cc|c|}\hline
Model &  \multicolumn{2}{c|}{Precision  \% } & \multicolumn{2}{c|}{Recall  \% }   &  \multicolumn{2}{c|}{F1  \% } & GPU time \\
&  train& val & train& val & train & val  & (epoch 80) \\\hline
DWM  &	85.9 & 83.6 &	74.4 & 73.8&	79.3 & 78.0	& 3h 9m \\
U18 & 89.5 & 87.6 &	87.0 & 83.4	& 88.1 & 85.3 &	3h 3m \\
U34 & 91.1 & 88.8 &	88.5 & 83.5 &	89.7 & 85.9	& 3h 45m \\
L18 & 90.3 & 88.4 &	87.3 & 82.0	& 88.7 & 84.9 &	2h 17m\\
L34 & 89.7 & 87.8	&87.4 & 83.5	& 88.5 & 85.4	& 3h 9m\\\hline
\end{tabular}
\caption{Summary of training results for {FCN} applied to $\sim 14,000$ 1-band panchromatic images}
\label{table::metricspan}
\end{centering}
\end{table}

The training times for all models are similar to those for the multispectral images, with Linknet again outperforming U-Net in speed and DWM matching the efficiency of L34.

Not surprisingly, all {FCN} perform somewhat worse for panchromatic images in terms of accuracy. Our 1-band version of DWM is the shallowest model and shows the poorest metrics. However, all the deeper models manage to achieve F1 scores of $\geq 85\%$.
 The fact that we only lose around 5\% in accuracy while using four times less information is remarkable. 
 
\begin{figure}[h!]
 \noindent\centerline{\includegraphics[width=\textwidth]{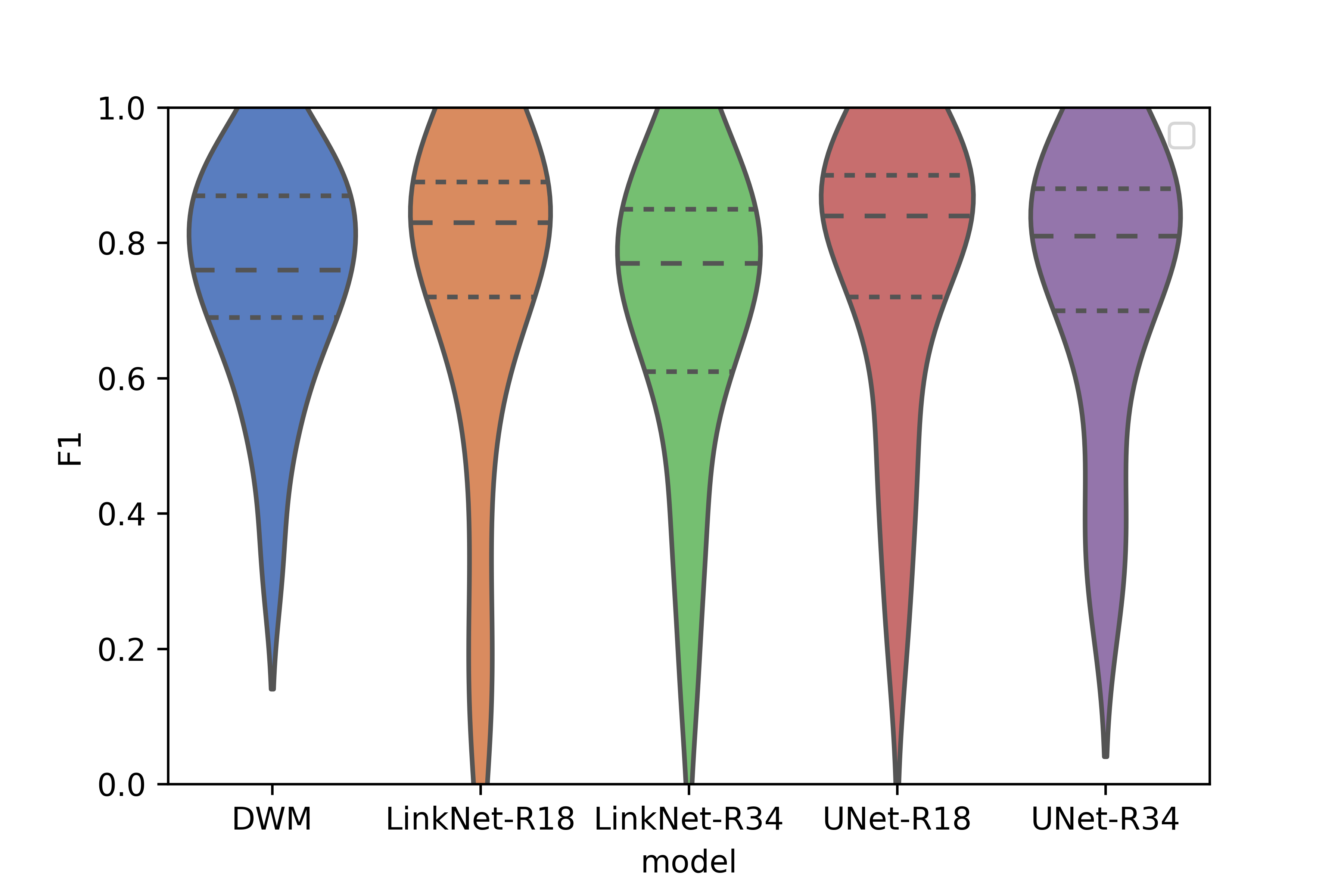}}
\caption{Violin plot for the F1 classification accuracy of all 5 \textbf{panchromatic} models for 17 independent full-size images. F1 histogram shown in color, short dashed lines show upper and lower quartiles and long-dashed line is the median.}
\label{fig::figx3}
\end{figure}

 { When we use the trained panchromatic models to classify our 17 hold-out full-size images, we see in Figure~\ref{fig::figx3} that the performance is somewhat degraded to an average F1 across the 5 models of $\sim 75\%$. The variance in these histograms is larger than for the multispectral images and the mean F1 scores are affected by a few poor masks. The \textit{median} F1 score across all models is $80\%$. L18 and U18 (the best panchromatic models) have median F1 scores of $\sim 85\%$ and a third of the images have F1 $>90\%$. Given that considerably more panchromatic imagery is available as compared to multispectral, it is feasible to simply discard poor classification results and still obtain sets of high accuracy river masks from panchromatic images.
 }
 
We are confident that these {FCN} far exceed the performance of any pixel-level classifier. As mentioned in the Introduction, the reason is that {FCN} not only learn highly nonlinear dependencies on pixel-level spectral information, but also large scale features such as the edges of river shorelines.  

 \subsection{(M)NDWI thresholding}
\label{ssec:ndwi}
To quantify the superiority of {FCN} over NDWI thresholding, we use the machine learning approach to optimize a single NDWI threshold across 80\% of our training data and then evaluate the accuracy of classification by using that threshold to make predictions for the remaining 20\% of labeled images. 

For both sets of training data we find a precision, recall, and F1 on the validation data of 52\%, 68\%, and 54\%, respectively, for an optimized threshold of 0.27. \cite{isikdogan2019seeing} finds somewhat better performance of a MNDWI on Landsat images with precision and recall of 57\% and 98\%, respectively. Many of the QB2, WV2 and WV3 images only have RGB-NIR bands so we are not able to test the performance of MNDWI.
As mentioned earlier,  recall is always higher than precision, but both are far less accurate than any deep learning model.

\section{Discussion}
\label{sec:discussion}

 \subsection{Quality of training data }
\label{ssec:trainingquality}
We want to emphasize that in most applications, including ours, no true ground-truth data are available at the pixel level. {To obtain such data, one would have to, for example, walk along river shorelines with a GPS tracker, which is not scalable. In manually generating training data, even for a human it can be hard to identify shorelines from satellite imagery, for example when there are tree shadows and wet or dry sandbanks. By using a LCCA to reduce noise in the training data, we may also exclude some smaller water tributaries that are not clearly connected to the main river. }

Most likely, at least a few percent of pixels are misclassified in the training data itself, which makes it all the more remarkable that the {FCN} are robust enough to achieve accuracies of $>92\%$ on multispectral validation data. 
The fact that the panchromatic model has a few percent lower accuracy may at least in part be due simply to a slightly worse quality of training data, e.g., due to small artifacts of matching re-sampled panchromatic images to labels derived from a multispectral counterpart. 

 \subsection{Model complexity }
\label{ssec:complexity}
\begin{figure}[h!]
 \noindent\centerline{\includegraphics[width=\textwidth]{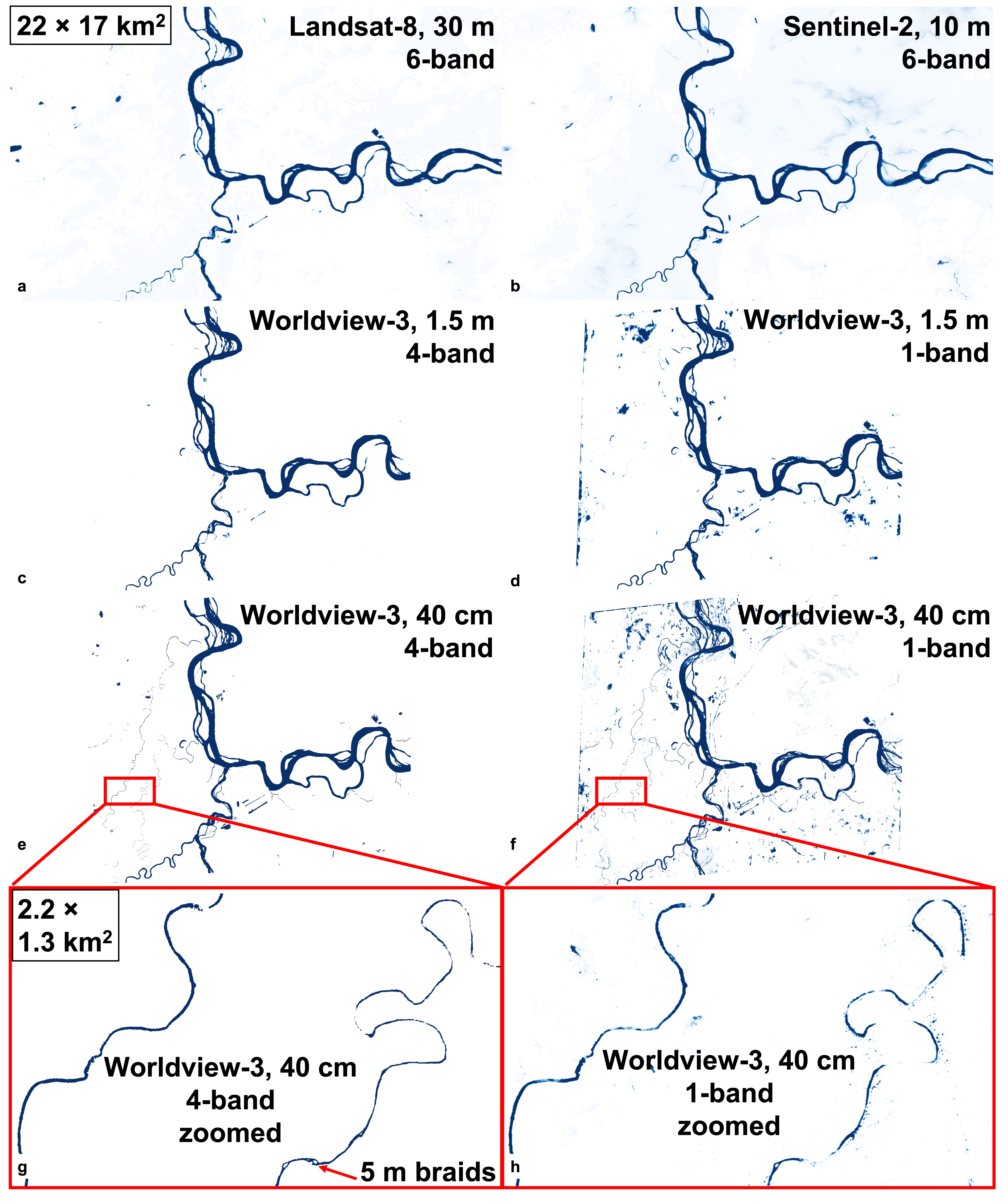}}
\caption{Comparison of classification results for the Tanana river at Nanana, AK for Landsat-8 at 30 m resolution (\textbf{a}), Sentinel-2 at 10 m (\textbf{b}), WV3 multispectral  (\textbf{c}) and panchromatic  (\textbf{d}) at 1.5 m and at pan-sharpened multispectral  (\textbf{e}) and native panchromatic  (\textbf{f}) resolutions of 40 cm. Zoomed-in panels show resolved 5 m wide river braids and islands  (\textbf{g}, \textbf{h}).
\label{fig::fig3}}
\end{figure}

Overfitting is a risk in machine learning when a model is overly complex for the (amount of) training data. Fortunately, we find that the training curves for all metrics converge for both 
 training and validation data. Visual inspection of masks on independent full-size images also confirms the excellent performance of our models (discussed further below in relation to Figures~\ref{fig::fig3}--\ref{fig::fig6}). Together, this alleviates concerns of overfitting, while the high accuracies suggest that the best models are sufficiently complex.  

\subsection{Post-processing}
\label{ssec:postprocessing}
Even the best FCN models may misclassify certain features with similar spectral signatures and shapes (e.g., sharp edges) as rivers. The panchromatic models are more sensitive to this because they use less information than available in multispectral data. Fortunately, most misclassified pixels can be removed by a simple post-processing step using the same techniques as discussed in the construction of our training data (e.g., a connected component analysis). {When we perform such post-processing, we also apply a hard threshold to obtain binary land-water classification labels. }
Examples are discussed in the next section.

\subsection{Comparison of classification results for different satellites, resolutions, and number of spectral bands }
\label{ssec:comparisons}
This section investigates and illustrates more details of the performance of our {FCN} for water remote sensing. 

Figure~\ref{fig::fig3} summarizes DWM classification results for a section of the Tanana river at Nanana, AK as observed by three different satellites and at different resolutions. Specifically, we consider a Landsat-8 image from July $1^\mathrm{st}$ 2021 at 30 m resolution, a Sentinel-2 image from October $9^\mathrm{th}$ 2020 with all bands sharpened to 10 m resolution, and both a panchromatic and an 8-band multispectral Worldview-3 image from July $6^\mathrm{th}$ 2019. For the latter we both down-sample the panchromatic image to the 1.5 m multispectral resolution and pan-sharpen the multispectral image to the 40 cm panchromatic resolution. In other words, these images cover two orders of magnitude in spatial resolution from 40 cm to 30 m. All images are cloud and snow free.

Classification of Landsat-8 and Sentinel-2 images is performed with the original 6-band DWM model and published training weights \citep{isikdogan2019seeing} and clearly detects the river body well (high recall) but also shows non-zero water probabilities for much of the background, which would have to be removed in post processing ({e.g., by a hard threshold}). 

Our 4-band DWM model applied to the native resolution multispectral WV3 image {qualitatively} shows excellent precision and recall (very few areas misclassified as water) without any post-processing. When we apply the model to a pan-sharpened multispectral image at 40 cm resolution, the model performs equally well and we can clearly resolve even smaller water features.

		 \begin{figure}[h!]
 \noindent\centerline{\includegraphics[width=\textwidth]{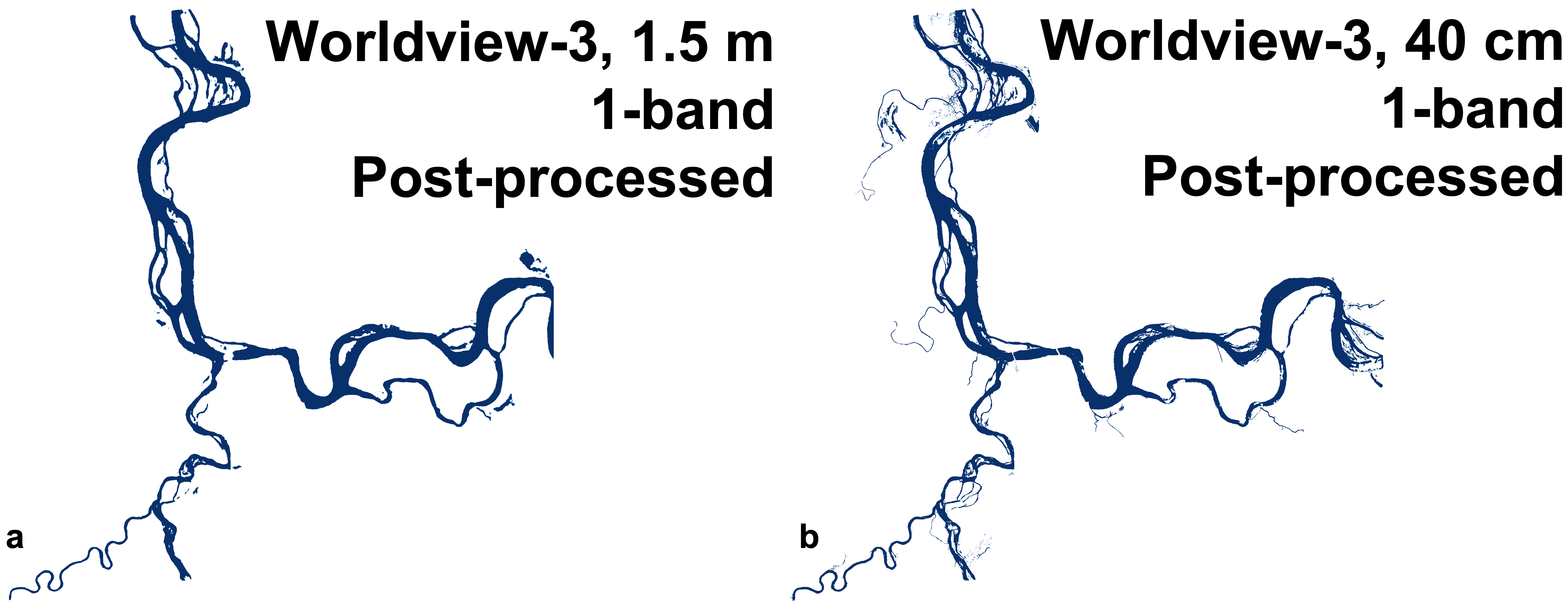}}
\caption{Panchromatic ({binary}) classification results from Figure~{\ref{fig::fig3}}e at 1.5 m (\textbf{a}) and f at 40 cm (\textbf{b}) resolutions after post-processing by, e.g., largest connected component operator.
\label{fig::fig2}}
\end{figure}

For the panchromatic images we apply our modified U18 {FCN}, which provides excellent recall on just the single band but has a lower precision (more noise) at both 1.5 m and 40 cm resolutions. Figure~\ref{fig::fig2} demonstrates how this noise can be removed by a post-processing step, achieving predictions almost as good as from the more information-rich multispectral data. 

Finally, we zoom in on one of the smaller tributaries where inference on both the (pan-sharpened) multispectral and panchromatic images can resolve braids and islands that are only a few meters across (Figure~\ref{fig::fig3}g--h). 

		 \begin{figure}[h!]
 \noindent\centerline{\includegraphics[width=\textwidth]{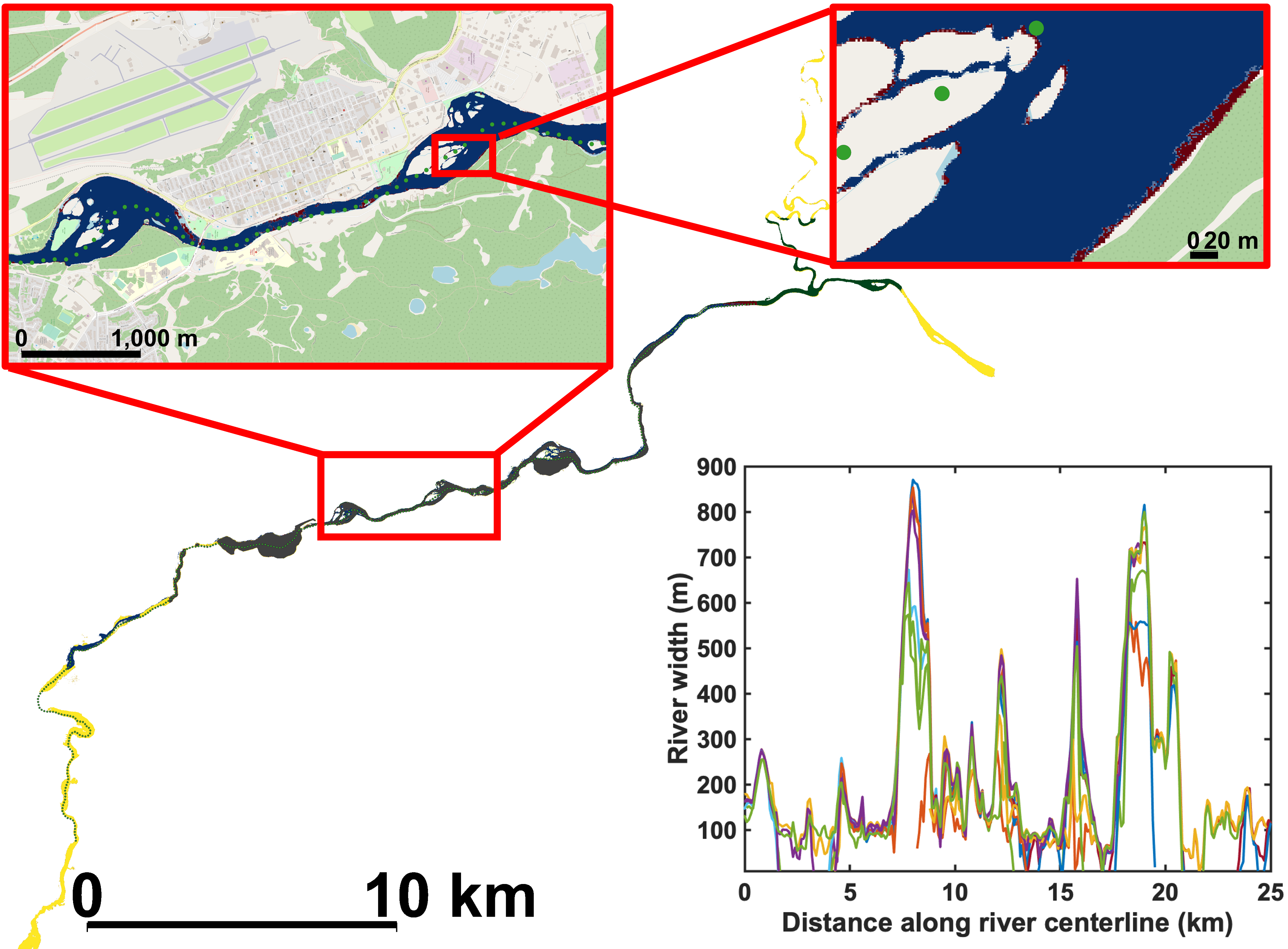}}
\caption{River classification for 16 years of satellite imagery covering the Yukon around Whitehorse, Canada. Smaller zoomed-in sections show masks for August 21$^\mathrm{st}$ 2016 and July 31$^\mathrm{st}$ 2017 down to meter-scale branches and width changes. In-set shows river widths projected onto 25 km of a centerline \citep{altenau2021surface}, which is shown as dots (spaced at 100 m intervals) overlaid on the water masks.  \copyright~OpenStreetMap contributors.
\label{fig::fig4}}
\end{figure}

{Not all applications in surface hydrology require meter-scale resolutions but some do, especially if one is interested in \textit{changes} in water bodies or other features over time.} 
As an example, one motivation of this work is to remotely sense river \textit{discharges} over time  \citep{feng2019comparing,brinkerhoff2020constraining,feng2021recent} using highly accurate river shorelines (widths) and water surface elevations (obtained by projecting our river masks onto a low-stage Digital Elevation Model, DEM), together with a flow law \citep{dai2018estimating}. For large rivers that can easily be resolved by, e.g., Landsat-8 or Sentinel-2, \textit{changes} in width of only a few (dozen) meters can correspond to major variations in discharge. Even meter-resolution images (and DEMs) are barely sufficient to track river surface elevations, slopes, and discharges with reasonable accuracy. 

As an illustration, Figure~\ref{fig::fig4} shows a stack of 18 water masks derived from multispectral images with DWM for a $\sim 650\ \mathrm{km}^2$ area covering the Yukon river around Whitehorse, Canada, for summer months between 2004 and 2020 (note that the coverages of individual satellites passes differ). The narrower parts of the river itself are $<100$ m but, more importantly, the \textit{changes} in river width over time are of the order of (tens of) meters, which can only be accurately resolved with meter-resolution imagery. This is illustrated by the zoomed-in insets that show river features down to meter scales as well as the differences in widths for masks at different times, and finally the derived river widths projected onto a 25~km long river centerline. 

		 \begin{figure}[h!]
 \noindent\centerline{\includegraphics[width=\textwidth]{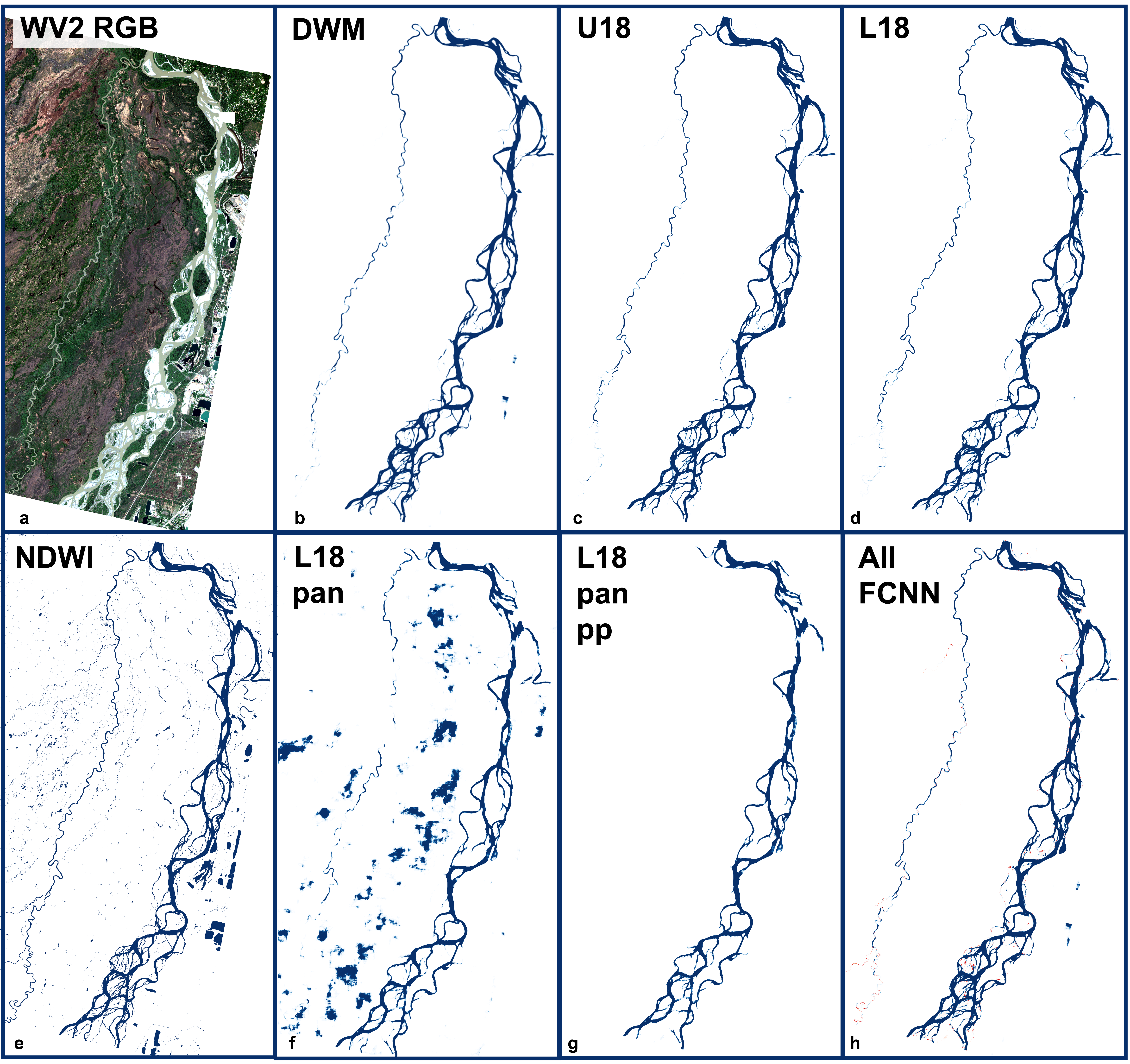}}
\caption{Water detection for $7,850\times 11,855$ pixel image (\textbf{a}) of the Tanana flowing through Fairbanks, AK (WV2, May 15$^\mathrm{th}$ 2015). Results shown for DWM (\textbf{b}), U18 (\textbf{c}), L18 (\textbf{d}), as well as a NDWI threshold of 0.27 for a multispectral image (\textbf{e}), and L18 for the panchromatic counterpart with (\textbf{f}) and without (\textbf{g}) post-processing. Last panel (\textbf{g}) shows stack of all {FCN}, each with a different color scheme, though only one is clearly visible due to the near-perfect overlap. Imagery \copyright~2015, MAXAR, Inc.
\label{fig::fig5}}
\end{figure}

\subsection{Comparison of classification results from different {FCN} }
\label{ssec:comparisonsfcnn}

\begin{figure}[h!]
 \noindent\centerline{\includegraphics[width=.8\textwidth]{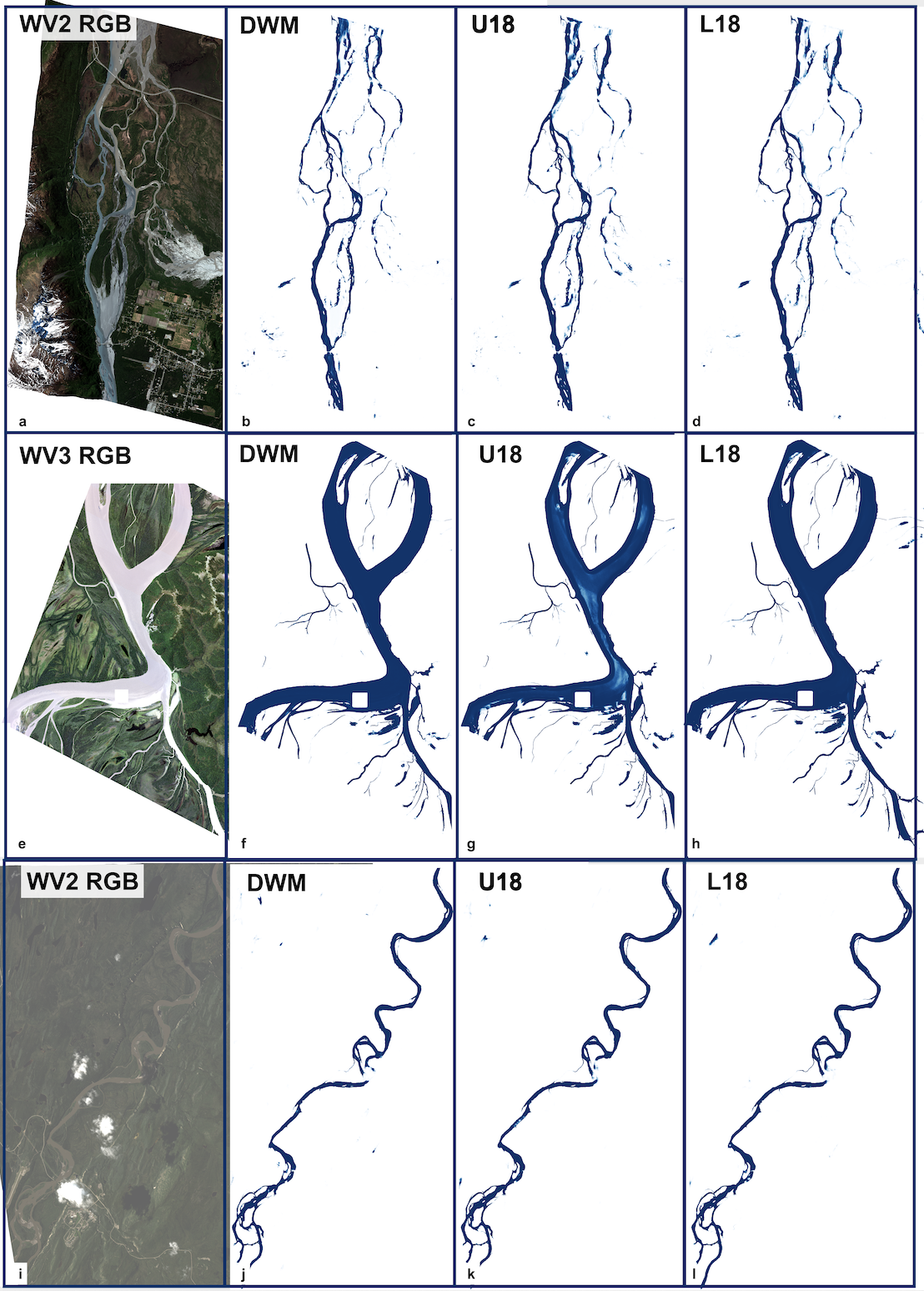}}
\caption{Water detection with (multispectral) DWM, U18, and L18 for the Knik near Palmer, Alaska ($11,237\times 14,795$ pxl, WV3, June 3$^\mathrm{rd}$ 2018; (\textbf{a}-\textbf{d})), the Yukon near St. Mary's, Alaska ($12,562\times 15,916$ WV3, August 30$^\mathrm{th}$ 2016; (\textbf{e}-\textbf{h})), and the Pelly at Faro, Canada ($8,879\times 8,952$ WV2, August 3$^\mathrm{rd}$ 2017; (\textbf{i}-\textbf{l})). Imagery \copyright~2021, MAXAR, Inc.
\label{fig::fig6}}
\end{figure}

Figures~\ref{fig::fig5} and \ref{fig::fig6} further illustrate the performance of different {FCN} in terms of classification on full-size multispectral and panchromatic imagery. 
Figures~\ref{fig::fig5} is a challenging scene for the Tanana as it flows through Fairbanks, Alaska, where the river exhibits complex braiding, wet and dry sandbanks, and roads and buildings that may easily be misclassified as water by other methods. Figures~\ref{fig::fig6} provides three more examples for the Knik and Yukon in Alaska, and the Pelly river in Canada. Together with the previously discussed classification results in Figures~\ref{fig::fig3} and \ref{fig::fig4}, these provide a broad test suite of different Arctic environments and river morphologies.

In terms of comparing the performance of the 10 presented {FCN}, both Linknet and U-Net with the Resnet-18 and Resnet-34 backbones give nearly identical classification results, so only results from the more efficient Resnet-18 versions are shown. In fact, classification results from \textit{all} the multispectral models are nearly indistinguishable. In the last panel of Figures~\ref{fig::fig5}, all {FCN} predictions are stacked together, each using a different color scheme. Besides a few buildings misclassified as water by DWM, there are no visible differences in the classification of the river itself between DWM, U18, U34, L18, and L34. The same is true for the other three examples in Figures~\ref{fig::fig6}. 

To be more quantitative, we select one model (which one does not matter) and compute F1 scores for all other models with respect to this reference. For all models and images in Figures~\ref{fig::fig5} and \ref{fig::fig6} the F1 scores are $>95\%$, i.e. less than 5\% of pixels are classified differently by different models.

Producing equally good masks from only panchromatic images is challenging. Linknet and U-Net show better performance than DWM in this case, and 1-band L18 classification results are shown in Figures~\ref{fig::fig5}. Clearly, the initial classification result has a lower precision but, again, the prediction is significantly improved by a connected-component analysis. By comparing to a traditional NDWI approach (with a threshold of 0.27 carefully optimized across all 14,000 training images), it appears that the 1-band L18 is less prone to misclassifying man-made structures as water. More generally, NDWI or other multi-band indices can of course not be used for panchromatic images, which makes these {FCN} an invaluable new tool. 

\begin{table}[h!]
\begin{centering}
\begin{tabular}{|c|c|c||c|c||c|c|}\hline
Model & Params.  & FLOPs  & CPU & RSS& CPU & RSS \\
& $\times 10^6$  & $\times 10^9$ & min:sec & GB & min:sec & GB  \\\hline
DWM &	 37.2 	& 95.0	& 32:52	& 48	& 30:26	& 40\\
U18& 	14.3 	& 43.7	& 19:47	& 213 & 17:51	& 191\\
U34	& 	24.4& 63.1	& 27:04	& 213& 24:34	& 191 \\ 
L18&	 11.5	& 22.7	& 11:23	&  117 & 10:40	&  103\\
L34& 21.6 	& 42.1	& 18:40	&  117 & 17:31	&  103 \\
\hline
\end{tabular}
\caption{Number of trainable parameters, floating-point-operations (FLOPs) for a $512 \times 512$ image, and inference CPU time and memory usage (maximum resident set size, RSS) on a single CPU thread for a $13,690\times 11,084=152 \times 10^6$ pixel WV3 image. The latter two columns are shown first for multispectral and then panchromatic classifications.}
\label{table::inferencemetrics}
\end{centering}
\end{table}

Table~\ref{table::inferencemetrics} provides a summary of quantitative performance metrics related to {classification} for all evaluated {FCN}, in addition to the {training} results in Tables~\ref{table::metrics} and \ref{table::metricspan}. Specifically, we provide the number of trainable parameters for each model, as well as the number of floating point operations (FLOPs) required to perform inference on a $512\times 512$ image. Both of these measures are determined by the number and type of convolutional layers in each model. Because those architectures are essentially the same for the multispectral and panchromatic versions, the numbers of trainable parameters and FLOPs for the multispectral model (shown in Table~\ref{table::inferencemetrics}) and panchromatic (not shown) are nearly the same. 
Table~\ref{table::inferencemetrics} also lists the CPU time and memory requirements for classification on a large WV3 image. For the most straightforward comparison, CPU times are shown for a single threaded execution on one 2.7GHz core. In practice, classification is performed on shared-memory multicore CPU and GPU nodes where all wall times are under 1.5 min. 

We can draw a number of conclusions from Table~\ref{table::inferencemetrics}. The most important is that while DWM is the most computationally expensive, it has by far the highest number of trainable parameters while simultaneously requiring significantly less memory. This is important and makes DWM the only FCN that can directly perform classification on full-size ($\sim 10^8$ pixel) satellite images on a reasonable consumer computer. The U-Net models are faster and showed the best accuracy metrics on training and validation data (though by an insignificant margin, Tables~\ref{table::metrics} and \ref{table::metricspan}). However, the memory requirements are prohibitive even on typical cluster nodes. 
The Linknet models still require $2.4\times$ more memory than DWM (versus $4.4\times $ for U-Net) and have $\sim 2$ -- $3\times$ fewer trainable parameters, but are up to $\sim 3\times$ faster than DWM. 

The multispectral and panchromatic versions of each model allow for different numbers of input layers but the main network architecture is the same for each panchromatic-multispectral {FCN} pair. As a result, the computation and memory requirements for the panchromatic classifications are within about 10\% from the multispectral ones. 

For reasons that are not entire clear, DWM does not perform as well for panchromatic images as the U-Net and Linknet models (Table~\ref{table::metricspan}). L18 offers a better combination of accuracy, computational cost and memory requirements. 

{\subsection{Detection of non-river water bodies}
The objective of this work was to classify \textit{rivers} rather than all water bodies. For this reason, we excluded small lakes, puddles, etc.~from our \textit{training} data with the goal of biasing our FCN to primarily detect river morphologies. Still, because the spectral signature of different types of water bodies can be relatively similar and the receptive field of our FCN is small relative to the spatial extent of rivers and lakes, our trained FCN will also classify most non-river water bodies (something we can eliminate by our LCCA post-processing if so desired). For research applications where small non-river water bodies are the primary interest, it may be worth re-training our models with slightly different training data that contain more of such features. The original DWM \citep{isikdogan2019seeing} was trained to detect all water pixels and its published training weights can be used to classify all water at $\gtrsim 10$ m resolutions. 
}

\subsection{Data access }
\label{ssec:access}
This work focusses on imagery covering the Alaskan and Canadian Arctic. Further model training would  benefit the application of these models globally. Unfortunately, the licensing of Maxar satellite imagery prevent the public sharing of imagery and thus labeled training data. However, we make all our models and training weights publicly available on github (\url{https://github.com/jmoortgat/DeepRiverFCN}). This will allow anyone with access to Maxar satellite imagery to directly apply our trained models for their purposes. Moreover, ‘transfer learning’ can be used to more efficiently (re-)train our models, starting from the shared training weights, on imagery in other regions or from other current and future satellites, such as Sentinel-2 and the Planet constellations.   

\section{Summary and Conclusions }
\label{ssec:conclusions}

This work advances the state-of-the-art in remote sensing of surface hydrological processes by taking advantage 1) of the highest (sub-) meter resolutions available in satellite imagery, {combined with} 2) the most advanced fully convolutional neural networks to classify water from {not only multispectral but also panchromatic data}.  
 
We evaluated the performance of a suite 10 {FCN} (5 for multispectral and 5 for panchromatic images) based on the most successful architectures from the literature in the last 5 years. 
We optimized the general purpose {FCN} for full-scale ($\sim 10^8$ pixels) satellite imagery and to allow for different numbers of spectral bands (1, 4, 6, and 8). To train the models, {we 
used a labor intensive process of NDWI thresholding, manual, and OpenCV image post-processing steps to generate 14,000 labeled multispectral training tiles. In a novel approach, we then used the same labels to train models on panchromatic images}, down-sampled to the multispectral resolution. Various data augmentation techniques, some specific to satellite imagery, were used to improve the model training.  

\textit{All} {FCN} algorithms achieved precision, recall, and F1 metrics of $>90\%$ on multispectral labeled validation data, while 4 models based on U-Net and Linknet demonstrated accuracy metrics of 
 $\gtrsim 85\%$ on panchromatic data. Classification on full-size images is both accurate and highly efficient ($<$ 1 min for all models on NVIDIA A40 GPU or RTX 3090).  
 
Memory requirements are a concern when deploying {FCN} for high resolution satellite imagery. Of the evaluated {FCN}, only DeepWaterMap \citep{isikdogan2019seeing} is sufficiently memory efficient to allow classification on full-size satellite imagery on consumer desktops or laptops. 

Altogether, the tools developed in this work, publicly shared on github,  allow the temporal tracking of changes in river morphologies with unprecedented accuracy and resolution. This opens up a wide range of opportunities to the remote sensing and surface hydrology communities.  

Even more broadly, these deep learning tools are application agnostic and were initially developed for general computer vision purposes. Our optimized codes for satellite imagery can be used to remotely sense a wide range of the Earth’s surface features other than water (forests, fires, faults, buildings, agricultural use, etc.), given a suitable set of training data.    

\section*{Acknowledgments}
The authors acknowledge funding from NASA Terrestrial Hydrology Program
 grant 80NSSC18K1497. Geospatial support for this work provided by the Polar Geospatial
 Center under NSF-OPP awards 1043681 and 1559691. DEMs provided by the Polar Geospatial
 Center under NSF-OPP awards 1043681, 1559691, and 1542736. 

\bibliographystyle{elsarticle-harv} 

\end{document}